\documentclass[12pt]{article}
\usepackage{amsmath,amssymb,amsfonts}

\newtheorem{thm}{Theorem}[section]
\newtheorem{prop}{Proposition}[section]

\def\rem{{\noindent{\bf Remarks.}\ }}

\def\sds{\strut \displaystyle}

\numberwithin{equation}{section}

\def\N{{\mathbb N}}
\def\R{{\mathbb R}}

\newtheorem{theorem}{Theorem}
\newtheorem{corollary}{Corollary}
\newtheorem{proposition}{Proposition}
\newtheorem{lemma}{Lemma}

\newtheorem{example}{Example}
\newtheorem{definition}{Definition}
\newcommand{\beq}{\begin{equation}}
\newcommand{\eeq}{\end{equation}}
\newcommand{\beas}{\begin{eqnarray*}}
\newcommand{\eeas}{\end{eqnarray*}}
\newcommand{\bea}{\begin{eqnarray}}
\newcommand{\eea}{\end{eqnarray}}
\newcommand{\bei}{\begin{itemize}}
\newcommand{\eei}{\end{itemize}}
\newcommand{\ben}{\begin{enumerate}}
\newcommand{\een}{\end{enumerate}}
\newcommand{\bet}{\begin{theorem}}
\newcommand{\eet}{\end{theorem}}
\newcommand{\bel}{\begin{lemma}}
\newcommand{\eel}{\end{lemma}}
\newcommand{\bep}{\begin{proposition}}
\newcommand{\eep}{\end{proposition}}
\newcommand{\bed}{\begin{definition}}
\newcommand{\eed}{\end{definition}}
\newcommand{\bec}{\begin{corollary}}
\newcommand{\eec}{\end{corollary}}
\newcommand{\bex}{\begin{example}}
\newcommand{\eex}{\end{example}}
\newcommand{\qed}{\quad\hbox{\vrule width 4pt height 6pt depth 1.5pt}}

\newcommand{\lam}{\lambda}

\newcommand{\hf}{{1 \over 2}}

\def\sb{{\cal B}}

\begin{document}

\markboth{{\small\it Recovery of Sparse Signals}} {{\small\it T.
Cai, G. Xu, and J. Zhang}}

\title{On Recovery of Sparse Signals via $\ell_1$ Minimization}

\author{T. Tony Cai\thanks{Department of Statistics, The Wharton School,
University of Pennsylvania, PA, USA;
e-mail: {\tt tcai@wharton.upenn.edu}.
Research supported in part by NSF Grant DMS-0604954.} ~
Guangwu Xu\thanks{Department of EE \& CS,
University of Wisconsin-Milwaukee, WI, USA; e-mail: {\tt
gxu4uwm@uwm.edu}} ~ and Jun Zhang\thanks{Department of EE \& CS,
University of Wisconsin-Milwaukee, WI, USA; e-mail: {\tt
junzhang@uwm.edu}}}

\date{}

\maketitle

\begin{abstract}
This article considers constrained $\ell_1$ minimization methods for
the recovery of high dimensional sparse signals in three settings:
noiseless, bounded error and Gaussian noise. A unified and
elementary treatment is given in these noise settings for two
$\ell_1$ minimization methods: the Dantzig selector and $\ell_1$
minimization with an $\ell_2$ constraint. The results of this paper
improve the existing results in the literature by weakening the
conditions and tightening the error bounds. The improvement on the
conditions shows that signals with larger support can be recovered
accurately. This paper also establishes connections between
restricted isometry property and the mutual incoherence property.
Some results of Candes, Romberg and Tao (2006) and Donoho, Elad, and
Temlyakov (2006) are extended.
\end{abstract}


\section{Introduction}
The problem of recovering a high-dimensional sparse signal based on a
small number of measurements, possibly corrupted by noise, has
attracted much recent attention. This problem arises in many different
settings, including model selection in linear regression, constructive
approximation,  inverse problems, and compressive sensing.

Suppose we have $n$ observations of the form
\beq
\label{model}
y = F\beta + z
\eeq
where the matrix $F\in \R^{n\times p}$ with $n\ll p$ is given and
$z\in \R^n$ is a vector of measurement errors.
The goal is to reconstruct the unknown vector $\beta\in \R^p$.
Depending on settings, the error vector $z$ can either be zero (in the
noiseless case), bounded, or
Gaussian where $z\sim N(0,\sigma^2I_n)$.  It is now well understood
that $\ell_1$ minimization provides an effective way for reconstructing a
sparse signal in all three settings.

A special case of particular interest is when no noise is present in
(\ref{model}) and $y=F\beta$. This is an underdetermined system of
linear equations with more variables than the number of equations.
It is clear that the problem is ill-posed and there are generally
infinite many solutions. However, in many applications the vector
$\beta$ is known to be sparse or nearly sparse in the sense that it
contains only a small number of nonzero entries. This sparsity
assumption fundamentally changes the problem, making unique solution
possible. Indeed in many cases the unique sparse solution can be
found exactly through $\ell_1$ minimization: \beq (P) \quad
\min\|\gamma \|_1 \quad \mbox{subject to} \quad F\gamma = y. \eeq
This $\ell_1$ minimization problem has been studied, for example, in
Fuchs \cite{Fuchs1}, Candes and Tao \cite{CanTao05} and Donoho
\cite{Donoho1}. Understanding the noiseless case is not only of
significant interest on its own right, it also provides deep insight
into the problem of reconstructing sparse signals in the noisy case.
See, for example, Candes and Tao \cite{CanTao05, CanTao07} and Donoho
\cite{Donoho1, Donoho2}.

When  noise is present, there are two well known $\ell_1$ minimization
methods. One is $\ell_1$ minimization under the $\ell_2$ constraint on the
residuals: \beq (P_1) \quad \min\|\gamma \|_1 \quad \mbox{subject
to} \quad \|y-F\gamma\|_2\le \epsilon. \eeq Writing in terms of the
Lagrangian function of ($P_1$), this is closely related to finding
the solution to the $\ell_1$ regularized least squares: \beq
\label{lasso} \min_\gamma \left\{\|y-F\gamma\|_2^2 + \rho\|\gamma
\|_1\right\}. \eeq The latter is often called the Lasso in the
statistics literature (Tibshirani \cite{Tib}). Tropp \cite{Tropp}
gave a detailed treatment of the $\ell_1$ regularized least squares
problem.

Another method, called the Dantzig selector, is recently proposed by
Candes and Tao \cite{CanTao07}. The Dantzig selector solves the
sparse recovery problem through $\ell_1$-minimization with a constraint
on the correlation between the residuals and the column vectors of $F$:
\begin{equation}\label{eq:1.3}
(DS)\quad  \min_{\gamma}
\|\gamma\|_1  \quad \mbox{subject to} \quad
\|F^T(y- F\gamma)\|_{\infty}\le \lambda.
\end{equation}
Candes and Tao \cite{CanTao07} showed that the Dantzig selector can be
computed by solving a linear program and it mimics the performance of
an oracle procedure up to a logarithmic factor $\log p$.

It is clear that regularity conditions are needed in order for these
problems to be well behaved. Over the last few years, many
interesting results for recovering sparse signals have been obtained
in the framework of the {\it Restricted Isometry Property} (RIP).
In their seminal work \cite{CanTao05,CanTao07}, Candes and Tao
considered sparse recovery problems in the RIP framework. They
provided beautiful solutions to the
problem under some conditions on the restricted isometry
constant  and restricted orthogonality constant (defined in Section
\ref{sec:notation}).
Several different conditions have been imposed in various settings.

In this paper, we consider $\ell_1$ minimization methods for the sparse
recovery problem in three cases: noiseless, bounded error and Gaussian
noise.  Both the Dantzig selector (DS) and $\ell_1$ minimization under
the $\ell_2$ constraint $(P_1)$ are considered. We give a
unified and elementary treatment for the two methods under the three
noise settings. Our results improve on the existing results in
\cite{Candes,CanRomTao,CanTao05,CanTao07} by weakening the
conditions and tightening the error bounds. In all cases we solve
the problems under the weaker condition
\[
\delta_{1.5k}+\theta_{k,1.5k}<1
\]
where $k$ is the sparsity index and $\delta$ and $\theta$ are
respectively the restricted isometry constant and restricted
orthogonality constant defined in Section \ref{sec:notation}.
The improvement on the condition shows that signals with larger
support can be recovered. Although our main interest is on recovering
sparse signals, we state
the results in the general setting of reconstructing an arbitrary signal.

Another widely used condition for sparse recovery is the
so called {\it Mutual Incoherence  Property} (MIP) which requires the
pairwise correlations among the column vectors of $F$ to be small.  See
\cite{DET,DonHuo,Fuchs1,Fuchs2,Tropp}.  We establish
connections between the concepts of RIP and MIP. As an application,
we present an improvement to a recent result of Donoho, Elad, and
Temlyakov  \cite{DET}.

The paper is organized as follows.
In Section \ref{sec:notation}, after basic notation and definitions
are reviewed, two elementary inequalities, which
allow us to make finer analysis of the sparse recovery problem, are introduced.
We begin the analysis of $\ell_1$ minimization methods for sparse recovery by
considering the exact recovery in the noiseless case in Section
\ref{sec:noiseless}. Our result improves the main result in Candes and
Tao
\cite{CanTao05} by using weaker conditions and providing tighter error
bounds. The analysis of the noiseless case provides insight to the
case when the observations are contaminated by noise. We then consider
the case of bounded error in Section \ref{sec:bounded.noise}. The
connections between the RIP and MIP are also explored.
The case of Gaussian noise is treated in Section
\ref{sec:Gaussian.noise}. The Appendix contains the proofs of some
technical results.

\section{Preliminaries}
\label{sec:notation}

In this section we first introduce basic notation and definitions,
and then develop some technical inequalities which will be used in
proving our main results.

Let $p\in \N$. Let $v=(v_1, v_2, \cdots, v_p) \in \R^p$ be a vector.
The support of $v$ is the subset of $\{1, 2, \cdots, p\}$ defined by
\[
\mbox{supp} (v) = \{ i : v_i\neq 0\}.
\]
For an integer $k\in \N$, a vector $v$ is said to be {\it
$k$-sparse} if $|\mbox{supp} (v)| \le k$.
For a given vector $v$ we shall denote by $v_{\max(k)}$ the
vector $v$ with all but the $k$-largest entries (in absolute value)
set to zero and
define $v_{-\max(k)} = v - v_{\max(k)}$, the vector $v$ with the
$k$-largest entries (in absolute value) set to zero.
We shall use the standard
notation $\|v\|_q$ to denote the $\ell_q$-norm of the vector $v$.

Let the matrix $F\in \R^{n\times p}$ and $1\le k \le p$, the {\it
$k$-restricted isometry constant} $\delta_k$ of $F$ is defined to be the
smallest constant such that
\begin{equation}\label{cond:2.1}
\sqrt{1-\delta_k}\|c\|_2 \le \|Fc\|_2 \le \sqrt{1+\delta_k}\|c\|_2
\end{equation}
for every vector $c$ which is $k$-sparse. If $k+k'\le p$, we can
define another quantity, the {\it $k, k'$-restricted orthogonality
constant} $\theta_{k, k'}$, as the smallest number that satisfies
\begin{equation}\label{cond:2.2}
|\langle Fc, Fc'\rangle| \le \theta_{k, k'}\|c\|_2\|c'\|_2,
\end{equation}
for all $c$ and $c'$ such that $c$ and $c'$ are $k$-sparse and
$k'$-sparse respectively, and have disjoint supports.
Candes and Tao \cite{CanTao05} showed that the constants $\delta_k$ and
$\theta_{k,k'}$ are related by the following inequalities,
\[
\theta_{k,k'} \le \delta_{k+k'}\le \theta_{k,k'} +\max(
\delta_{k},\delta_{k'}).
\]

Another useful property is as follows.
\begin{prop}\label{prop:2.1}
If $k+\sum_{i=1}^l k_i\le p$, then
\[
\theta_{k, \sum_{i=1}^l k_i}\le\sqrt{ \sum_{i=1}^l \theta_{k, k_i}^2}.
\]
In particular, $\theta_{k, \sum_{i=1}^l k_i}\le
\sqrt{ \sum_{i=1}^l \delta_{k+k_i}^2}$.
\end{prop}

\noindent
{\bf Proof of Proposition~\ref{prop:2.1}.} Let $c$ be $k$-sparse and
$c'$ be $(\sum_{i=1}^l k_i)$-sparse. Suppose their supports are disjoint.
Decompose $c'$ as
\[
c'=c'_1+c'_2+\cdots + c'_l
\]
such that $c'_i$ is $k_i$-sparse for $i = 1, \cdots, j$ and
$\mbox{supp}(c')_i \cap \mbox{supp}(c')_j=\emptyset$ for $i\neq j$.
We have
\begin{eqnarray*}
|\langle Fc, Fc'\rangle| &=&|\langle Fc, \sum_{i=1}^l Fc'_i\rangle|
\le \sum_{i=1}^l |\langle Fc, Fc'_i \rangle|\\
&\le & \sum_{i=1}^l \theta_{k, k_i}\|c\|_2\|c'_i\|_2 =\|c\|_2 \sqrt{
\sum_{i=1}^l \theta_{k, k_i}^2}
\sqrt{ \sum_{i=1}^l\|c'_i\|_2^2}\\
&=&\sqrt{ \sum_{i=1}^l\theta_{k, k_i}^2}\|c\|_2\|c'\|_2.
\end{eqnarray*}
This yields $\theta_{k,\sum_{i=1}^l k_i}\le
\sqrt{\sum_{i=1}^l \theta_{k, k_i}^2}$.
Since $\theta_{k, k'} \le \delta_{k+k'}$, we also have
$\theta_{k, \sum_{i=1}^l k_i}\le
\sqrt{ \sum_{i=1}^l \delta_{k+k_i}^2}$. \qed

\medskip\noindent
{\bf Remark:} Different conditions on $\delta$ and $\theta$  have
been used in the literature. For example, Candes and Tao
\cite{CanTao07} imposes $\delta_{2k}+\theta_{k,2k}<1$ and Candes
\cite{Candes} uses $\delta_{2k}<\sqrt{2}-1$. A direct consequence of
Proposition \ref{prop:2.1} is that $\delta_{2k}< \sqrt{2} -1$ is in
fact a strictly stronger condition than $\delta_{2k} + \theta_{k,2k}
< 1$ since Proposition \ref{prop:2.1} yields $\theta_{k,2k} \le
\sqrt{\delta_{2k}^2 + \delta_{2k}^2} = \sqrt{2}\delta_{2k}$ which
means that $\delta_{2k}< \sqrt{2} -1$ implies
$\delta_{2k}+\theta_{k,2k}<1$.

\medskip
We now introduce two useful elementary inequalities. These
inequalities allow us to perform finer estimation on $\ell_1,l_2$ norms.

\begin{prop}\label{prop:2.2}
Let $w$ be a positive integer. For any descending chain of real
numbers
\[
a_1\ge a_2 \ge \cdots \ge a_w\ge a_{w+1}\ge \cdots \ge a_{2w}\ge 0,
\]
we have
\[
\sqrt{a_{w+1}^2+a_{w+2}^2+\cdots +a_{2w}^2} \le
\frac{a_1+a_2+\cdots+a_w+a_{w+1}+\cdots+a_{2w}}{2\sqrt{w}}.
\]
\end{prop}

\noindent {\bf Proof of Proposition \ref{prop:2.2}}. Since $a_i\ge
a_j$ for $i<j$, we have
\begin{eqnarray*}
(a_1+a_2+\cdots+a_{2w})^2 &=&a_1^2+a_2^2+\cdots a_{2w}^2+2\sum_{i<j}a_ia_j\\
      &\ge &a_1^2+a_2^2+\cdots a_{2w}^2+2\sum_{i<j}a_j^2\\
      &=& a_1^2+3a_2^2+\cdots +(2w-1)a_w^2+\\
      &&    +  (2w+1)a_{w+1}^2+\cdots+(4w-3)a_{2w-1}^2+(4w-1)a_{2w}^2\\
      &=& \big(a_1^2+(4w-1)a_{2w}^2\big)+\big(3a_2^2+(4w-3)a_{2w-1}^2\big)+
                 \cdots\\
      &&    +\big((2w-1)a_w^2+(2w+1)a_{w+1}^2\big)\\
      &\ge & 4w a_{2w}^2+4w a_{2w-1}^2+\cdots 4w a_{w+1}^2. \qed
\end{eqnarray*}

Proposition~\ref{prop:2.2} can be used to improve the main result in
Candes and Tao \cite{CanTao07} by weakening the condition to
$\delta_{1.75k}+\theta_{k, 1.75k}<1$. However, the next proposition,
which we will use in proving our main results, is more powerful for
our applications.

\begin{prop}\label{prop:2.3}
Let $w$ be a positive integer. Then any descending chain of real
numbers
\[
a_1\ge a_2 \ge \cdots \ge a_w\ge a_{w+1}\ge \cdots \ge a_{3w}\ge 0
\]
satisfies
\[
\sqrt{a_{w+1}^2+a_{w+2}^2+\cdots +a_{3w}^2} \le
\frac{a_1+\cdots+a_w+2(a_{w+1}+\cdots+a_{2w}) +
a_{2w+1}+\cdots+a_{3w}} {2\sqrt{2w}}.
\]
\end{prop}
The proof of Proposition \ref{prop:2.3} is given in the Appendix.

\section{Signal Recovery in the Noiseless Case}
\label{sec:noiseless}

As mentioned in the introduction we shall consider recovery of
sparse signals in three cases: noiseless,  bounded error, and
Gaussian noise. We begin in this section by considering the problem
of exact recovery of sparse signals when no noise is present. This
is an interesting problem by itself and has been considered in a
number of papers. See, for example,  Fuchs \cite{Fuchs1}, Donoho
\cite{Donoho1}, and Candes and Tao \cite{CanTao05}. More importantly,
the solutions to this ``clean'' problem shed light on the noisy
case. Our result improves the main result given in Candes and Tao
\cite{CanTao05}. The improvement is obtained by using the technical
inequalities we developed in previous section. Although the focus is
on recovering sparse signals, our results are stated in the general
setting of reconstructing an arbitrary signal.

Let $F\in \R^{n\times p}$ with $n<p$ and suppose we are given $F$ and
$y$ where $y=F\beta$ for some unknown  vector $\beta$. The goal
is to recover $\beta$ exactly when it is sparse.
Candes and Tao  \cite{CanTao05} showed that a sparse solution
can be obtained by $\ell_1$ minimization which is then solved via linear
programming.
\begin{thm}[Candes and Tao \cite{CanTao05}]
\label{thm:3.1} Let $F\in \R^{n\times p}$. Suppose $k\ge 1$
satisfies
\begin{equation}\label{cond:2.4}
\delta_k + \theta_{k,k}+\theta_{k, 2k} < 1.
\end{equation}
Let $\beta$ be a $k$-sparse vector and $y:=F\beta$. Then $\beta$ is the
unique minimizer to the problem
\[
(P) \quad \min\|\gamma \|_1 \quad \mbox{subject to} \quad F\gamma = y.
\]
\end{thm}
We shall show that this result can be further improved by a
transparent argument.
A direct application of Proposition \ref{prop:2.3} yields the
following result which improves Theorem \ref{thm:3.1}.
 by weakening the condition from
\[
\delta_k + \theta_{k,k}+\theta_{k, 2k} < 1,
\]
to
\[
\delta_{1.5k}+\theta_{k, 1.5k}<1.
\]

\begin{thm}\label{thm:3.2}
Let $F\in \R^{n\times p}$. Suppose $k\ge 1$ satisfies
\[
\delta_{1.5k}+\theta_{k, 1.5k} < 1
\]
and $y=F\beta$.  Then the minimizer $\hat \beta$ to the problem
\[
(P) \quad \min\|\gamma \|_1 \quad \mbox{subject to} \quad F\gamma = y
\]
obeys
\[
\|\hat \beta - \beta\|_2 \le C_0 k^{-\hf} \|\beta_{-\max(k)}\|_1
\]
where $C_0= {2\sqrt{2}(1-\delta_{1.5k})\over 1 - \delta_{1.5k} -
  \theta_{k,1.5k}}$.

In particular, if $\beta$ is a $k$-sparse vector, then
$\hat \beta = \beta$, i.e., the $\ell_1$ minimization recovers $\beta$
exactly.
\end{thm}

\medskip\noindent
{\bf Proof of Theorem~\ref{thm:3.2}:} The proof relies on
Proposition  \ref{prop:2.3} and makes use of
the ideas from \cite{CanRomTao,CanTao05,CanTao07}.
In this proof, we shall also identify a vector $v=(v_1, v_2,\cdots,
v_p)\in \R^p$  as a function $v: \{1,2,\cdots, p\}\rightarrow \R$ by
assigning $v(i)=v_i$.

Let $\hat \beta$ be a solution to the $\ell_1$ minimization problem
(P).
Let $T_0=\{n_1, n_2, \cdots, n_k\}\subset \{1,2,\cdots, p\}$ be the
support of $\beta_{\max(k)}$ and let $h = \hat{\beta}-\beta$.  Write
\[
\{1,2,\cdots, p\}\setminus \{n_1, n_2, \cdots, n_k\} = \{n_{k+1},
n_{k+2}, \cdots, n_p\}
\]
such that $|h(n_{k+1})| \ge |h(n_{k+2})| \ge |h(n_{k+3})| \ge \cdots.$
Fix an integer $t > 0$  and let
\[
T_1=\{n_{k+1}, n_{k+2}, \cdots, n_{(t+1)k}\}, \mbox{ }
T_2=\{n_{(t+1)k+1}, n_{(t+1)k+2}, \cdots, n_{(2t+1)k}\}, \cdots.
\]

For a subset $E\subset \{1,2,\cdots, m\}$, we use $I_E$ to denote
the characteristic function of $E$, i.e.,
\[
I_E(j) =\left\{ \begin{array}{ll} 1 & \mbox{ if } j\in E,\\
                                  0 & \mbox{ if } j\notin E.\\
                                  \end{array}\right.
 \]
For each $i$, let $h_i = hI_{T_i}.$
Then $h$ is decomposed to $h = h_0+h_1+h_2+\cdots$. Note that $T_i$'s
are pairwise disjoint, ${\rm supp}(h_i)\subset T_i$, and $|T_0|=k,
|T_i| = tk$ for $i>0$. Without loss of generality, we assume $k$ is
divisible by $4$.

For each $i>1$, we divide $h_i$ into two halves in the following
manner
\[
h_i = h_{i1}+h_{i2} \mbox{ with } h_{i1} = h_iI_{T_{i1}}, \mbox{ and
} h_{i2} = h_iI_{T_{i2}},
\]
where $T_{i1}$ is the first half of $T_i$, i.e.,
\[
T_{i1}=\{n_{((i-1)t+1)k+1}, n_{((i-1)t+1)k+2},\cdots,
n_{((i-1)t+1)k+\frac{k}2}\} ,
\]
and $T_{i2}=T_i\setminus T_{i1}$.

 We
shall treat $h_1$ as a sum of four functions and divide $T_1$ into
$4$ equal parts  $T_1=T_{11}\cup T_{12}\cup T_{13}\cup T_{14}$ with
\[
T_{11}=\{n_{k+1},n_{k+2},\cdots,n_{k+t\frac{k}4}\}, \mbox{ }
 T_{12}=\{n_{k+t\frac{k}4+1},\cdots,n_{k+t\frac{k}2}\},
 \]
 \[
T_{13}=\{n_{k+t\frac{k}2+1},\cdots,n_{k+t\frac{3k}4}\} \mbox{  and }
T_{14}=\{n_{k+t\frac{3k}4+1},\cdots,n_{k+tk}\}.
\]
We then define $h_{1i}$ for $1\le i \le 4$ by $h_{1i}(j) =
h_1I_{T_{1i}}$. It is clear that $\sds h_1=\sum_{i=1}^4h_{1i}$.

Note that
\begin{equation}\label{ineq:3.1}
 \sum_{i\ge 1}\|h_i\|_1 \le \|h_0\|_1 + 2 \|\beta_{-\max(k)}\|_1.
\end{equation}
In fact, since $\|\beta\|_1 \ge \|\hat{\beta}\|_1$, we have
\beas
\|\beta\|_1 &\ge& \|\hat{\beta}\|_1 = \|\beta+h\|_1 =\|\beta_{\max(k)}+h_0\|_1+
\|h-h_0+\beta_{-\max(k)}\|_1 \\
&\ge& \|\beta_{\max(k)}\|_1-\|h_0\|_1 + \sum_{i\ge 1} \|h_i\|_1 - \|\beta_{-\max(k)}\|_1.
\eeas
Since $\|\beta\|_1= \|\beta_{\max(k)}\|_1+ \|\beta_{-\max(k)}\|_1$, this yields
$\sum_{i\ge 1} \|h_i\|_1 \le \|h_0\|_1 + 2 \|\beta_{-\max(k)}\|_1.$

The following claim follows from our Proposition~\ref{prop:2.3}.

{\bf Claim}
\begin{equation}\label{ineq:3.2}
\|h_{13}+h_{14}\|_2+\sum_{i\ge 2}\|h_i\|_2 \le
\frac{\sum_{i\ge 1} \|h_i\|_1}{\sqrt{tk}} \le
\frac{\|h_0\|_2}{\sqrt{t}} + \frac{2 \|\beta_{-\max(k)}\|_1}{\sqrt{tk}}.
\end{equation}
In fact, from Proposition~\ref{prop:2.3} and the fact that $\|h_{11}\|_1\ge \|h_{12}\|_1\ge\|h_{13}\|_1\ge\|h_{14}\|_1$, we have
\[
\|h_{12}\|_1+2\|h_{13}\|_1+\|h_{14}\|_1\le
\frac{2}3\big(2\|h_{11}\|_1+2\|h_{12}\|_1+\|h_{13}\|_1+\|h_{14}\|_1\big).
\]
It then follows from Proposition~\ref{prop:2.3} that
\begin{eqnarray*}
\|h_{13}+h_{14}\|_2
&\le&\frac{\|h_{12}\|_1+2\|h_{13}\|_1+\|h_{14}\|_1}{2\sqrt{\frac{tk}2}}\\
&\le&\frac{2}3
\frac{2\|h_{11}\|_1+2\|h_{12}\|_1+\|h_{13}\|_1+\|h_{14}\|_1}{2\sqrt{\frac{tk}2}}\\
&\le&
\frac{2\|h_{11}\|_1+2\|h_{12}\|_1+\|h_{13}\|_1+\|h_{14}\|_1}{2\sqrt{tk}}.
\end{eqnarray*}
Proposition~\ref{prop:2.3} also yields
\[
\|h_2\|_2 \le
\frac{\|h_{13}+h_{14}\|_1+2\|h_{21}\|_1+\|h_{22}\|_1}{2\sqrt{tk}}
\]
and
\[
\|h_i\|_2 \le \frac{ \|h_{(i-1)2}\|_1+2\|h_{i1}\|_1+\|h_{i2}\|_1
  }{2\sqrt{tk}}
\]
for any $i>2$. Therefore,
\begin{eqnarray*}
\|h_{13}+h_{14}\|_2+\sum_{i\ge 2}\|h_i\|_2 &\le&
\frac{2\|h_{11}\|_1+2\|h_{12}\|_1+\|h_{13}\|_1+\|h_{14}\|_1}{2\sqrt{tk}}\\
& & +\frac{\|h_{13}+h_{14}\|_1+2\|h_{21}\|_1+\|h_{22}\|_1}{2\sqrt{tk}}\\
&&+\frac{\|h_{22}\|_1+2\|h_{31}\|_1+\|h_{32}\|_1 }{2\sqrt{tk}}+\cdots \\
&\le& \frac{2\|h_1\|_1+2\|h_2\|_1+2\|h_3\|_1+\cdots }{2\sqrt{tk}}\\
&=&
\frac{\sum_{i\ge 1} \|h_i\|_1}{\sqrt{tk}}\\
&\stackrel{\mbox{by (\ref{ineq:3.1})} }{\le}&
\frac{\|h_0\|_1 + 2 \|\beta_{-\max(k)}\|_1}{\sqrt{tk}}
\le
\frac{\|h_0\|_2}{\sqrt{t}} + \frac{2 \|\beta_{-\max(k)}\|_1}{\sqrt{tk}}.
\end{eqnarray*}

In the rest of our proof we write $h_{11}+h_{12}=h_1'$. Note that
$Fh = F\hat \beta - F\beta = 0$. So
\begin{eqnarray*}
0&=& |\langle Fh, F(h_0+h_1')\rangle|\\
 & =   & |\langle F(h_0+h_1'),F(h_0+h_1')\rangle + \langle
F(h_{13}+h_{14}),F(h_0+h_1')\rangle
    +\sum_{i\ge 2}\langle Fh_i, F(h_0+h_1')\rangle | \\
  & \stackrel{(\ref{cond:2.1},\ref{cond:2.2})}{\ge} &
(1-\delta_{(\frac{1}2t+1)k}) \|h_0+h_1'\|_2^2 -
\theta_{\frac{1}2tk,(\frac{1}2t+1)k}\|h_{13}+h_{14}\|_2\|h_0+h_1'\|_2\\
      &&-\sum_{i\ge 2}\theta_{tk,(\frac{1}2t+1)k}\|h_i\|_2\|h_0+h_1'\|_2 \\
  & \ge & \|h_0+h_1'\|_2 \bigg((1-\delta_{(\frac{1}2t+1)k})\|h_0+h_1'\|_2 -
\theta_{tk,(\frac{1}2t+1)k}\big(\|h_{13}+h_{14}\|_2+\sum_{i\ge 2}\|h_i\|_2\big)\bigg) \\
&\stackrel{(\ref{ineq:3.2})}{\ge}& \|h_0+h_1'\|_2
\bigg((1-\delta_{(\frac{1}2t+1)k}) \|h_0+h_1'\|_2 -
\theta_{tk,(\frac{1}2t+1)k}\frac{\|h_0\|_2}{\sqrt{t}}
- \theta_{tk,(\frac{1}2t+1)k}\frac{2 \|\beta_{-\max(k)}\|_1}{\sqrt{tk}}
\bigg)\\
&\ge & \|h_0+h_1'\|_2\left\{ \bigg(1-\delta_{(\frac{1}2t+1)k}-
            \frac{\theta_{tk,(\frac{1}2t+1)k}}{\sqrt{t}}\bigg)\|h_0+h_1'\|_2 - \theta_{tk,(\frac{1}2t+1)k}\frac{2 \|\beta_{-\max(k)}\|_1}{\sqrt{tk}}\right\}.
\end{eqnarray*}
Take $t=1$. Then
\[
\|h_0+h_1'\|_2\le {2 \theta_{k,1.5k}\over 1 - \delta_{1.5k} -
  \theta_{k,1.5k}} k^{-\hf} \|\beta_{-\max(k)}\|_1
\]
It then follows from (\ref{ineq:3.2}) that
\beas
\|h\|_2^2 &=&\|h_0+h_1'\|_2^2 + \|h_{13}+h_{14}\|_2^2+\sum_{i\ge 2}\|h_i\|_2^2
\le\|h_0+h_1'\|_2^2 + (\|h_{13}+h_{14}\|_2+\sum_{i\ge 2}\|h_i\|_2)^2 \\
&\le&2(\|h_0+h_1'\|_2 + 2k^{-\hf} \|\beta_{-\max(k)}\|_1)^2
\le 2\left({2(1-\delta_{1.5k})\over 1 - \delta_{1.5k} -
  \theta_{k,1.5k}} k^{-\hf} \|\beta_{-\max(k)}\|_1\right)^2. \qed
\eeas

\rem
\begin{enumerate}
\item Candes and Tao \cite{CanTao07} considers the Gaussian noise case. A
special case with noise level $\sigma=0$ of Theorem 1.1 in that
paper improves Theorem \ref{thm:3.1} by weakening the condition from
$\delta_k + \theta_{k,k}+\theta_{k, 2k} < 1$ to
$\delta_{2k}+\theta_{k, 2k}<1.$

\item This theorem improves the results in
\cite{CanTao05,CanTao07}. The condition $\delta_{1.5k}+\theta_{k,
1.5k} < 1$ is weaker than $\delta_{k}+\theta_{k,k}+\theta_{k, 2k} <
1$ and $\delta_{2k}+\theta_{k, 2k} < 1$.
\item Note that the condition $\delta_{1.75k}< \sqrt{2}-1$
implies $\delta_{1.5k}+\theta_{k, 1.5k} < 1$. This is due to the
fact $\delta_{1.5k}+\theta_{k, 1.5k}\le
\delta_{1.5k}+\sqrt{\delta_{1.75k}^2+ \delta_{1.75k}^2}\le
(\sqrt{2}+1)\delta_{1.75k}$ by Proposition~\ref{prop:2.1}. The
condition $\delta_{1.5k} + \delta_{2.5k}<1$, which involves only
$\delta$, can also be used.

\item The quantity $t$ in the proof can be any number such that $tk\in \N$.
As pointed out in \cite{CanTao05,CanTao07}, other values of $t$ may
be used for obtaining some interesting results.
\end{enumerate}

\section{Recovery of Sparse Signals in Bounded Error}
\label{sec:bounded.noise}

We now turn to the case of bounded error. The results obtained in this
setting have direct implication for the case of Gaussian
noise which will be discussed in Section \ref{sec:Gaussian.noise}.

Let $F\in \R^{n\times p}$ and let
\[
y = F\beta + z
\]
where the noise $z$ is bounded, i.e., $z\in \sb$ for some bounded set
$\sb$. In this case the noise $z$ can either be stochastic or
deterministic. The $\ell_1$ minimization approach is to estimate $\beta$
by the minimizer $\hat \beta$ of
\[
\min\|\gamma\|_1 \quad \mbox{subject to} \quad y-F\gamma \in \sb.
\]
We shall specifically consider two cases: $\sb=\{z:\;
\|F^Tz\|_{\infty}\le \lambda\}$ and $\sb=\{z:\; \|z\|_2 \le
\epsilon\}$.
Our results improve the results in Candes and Tao
\cite{CanTao05,CanTao07} and Donoho, Elad and Temlyakov \cite{DET}.

We shall first consider
\[
y = F\beta + z \quad \mbox { where $z$ satisfies }\quad
\|F^Tz\|_{\infty}\le \lambda.
\]
Let $\hat{\beta}$ be the solution to the (DS) problem, i.e.,
$\hat{\beta}$ is obtained by solving
\begin{equation}
\label{DS.Bounded}
 \min_{\gamma \in \R^p}
\|\gamma\|_1  \mbox{ \hspace{5mm} subject to \hspace{5mm} }
\|F^T\big( y - F\gamma \big)\|_{\infty}\le \lambda.
\end{equation}
The Dantzig selector $\hat \beta$ has the following property.
\begin{thm}\label{thm:4.1}
Suppose $\beta\in \R^{p}$ and
$y=F\beta +z$ with $z$ satisfying $\|F^Tz\|_{\infty}\le
\lambda$. If
\begin{equation}\label{cond:4.1}
\delta_{1.5k}+\theta_{k, 1.5k} < 1,
\end{equation}
then the solution $\hat{\beta}$ to (\ref{DS.Bounded}) obeys
\begin{equation}
\| \hat{\beta}-\beta\|_2\le C_1 k^{\frac{1}2}\lambda
+C_2k^{-\frac{1}2} \|\beta_{-\max(k)}\|_1
\end{equation}
with $C_1=\frac{2\sqrt{3}}{1-\delta_{1.5k}-\theta_{k, 1.5k}}$, and
$C_2=\frac{2\sqrt{2}(1-\delta_{1.5k})}{1-\delta_{1.5k}-\theta_{k,
1.5k}}$.

In particular, if $\beta$ is a $k$-sparse vector, then $\|
\hat{\beta}-\beta\|_2\le C_1 k^{\frac{1}2}\lambda$.
\end{thm}

\noindent {\bf Proof of Theorem~\ref{thm:4.1} }. We shall use the
same notation as in the proof of Theorem~\ref{thm:3.2}. Since
$\|\beta\|_1\ge\|\hat{\beta}\|_1$, letting $h=\hat{\beta}-\beta$ and
following essentially the same steps as in the first part of the
proof of Theorem~\ref{thm:3.2}, we get
\[
|\langle Fh, F(h_0+h_1')\rangle|\ge \|h_0+h_1'\|_2 \left\{
\bigg(1-\delta_{1.5k}-\theta_{k,1.5k}\bigg)\|h_0+h_1'\|_2 -
\theta_{k,1.5k}\frac{2 \|\beta_{-\max(k)}\|_1}{\sqrt{k}} \right\}.
\]

If $\|h_0+h_1'\|_2=0$, then $h_0=0$ and $h_1'=0$. The latter forces
that $h_j=0$ for every $j >1$, and we have $\hat{\beta}-\beta=0$.
Otherwise
\[
\|h_0+h_1'\|_2\le \frac{|\langle Fh, F(h_0+h_1')\rangle
|}{\big(1-\delta_{1.5k}-\theta_{k,1.5k}\big)\|h_0+h_1'\|_2}+
\frac{2\theta_{k,1.5k}\|\beta_{-\max(k)}\|_1}{\big(1-\delta_{1.5k}-\theta_{k,1.5k}\big)\sqrt{k}}.
\]
To finish the proof, we observe the following.
\begin{enumerate}
\item $|\langle Fh, F(h_0+h_1')\rangle |\le \sqrt{1.5k}\, 2\lambda \|h_0+h_1'\|_2$.

In fact,
let $F_{T_0\cup T_{10}\cup T_{11}}$ be the $n\times (1.5k)$ submatrix
obtained by extracting the columns of $F$ according to the indices
in $T_0\cup T_{10}\cup T_{11}$, as in \cite{CanTao07}. Then
\begin{eqnarray*}
|\langle Fh, F(h_0+h_1')\rangle|&=&|\langle (F\hat{\beta}-y)+z,
F_{T_0\cup T_{10}\cup T_{11}}(h_0+h_1')\rangle |\\
&=&|\langle F_{T_0\cup T_{10}\cup T_{11}}^T\big((F\hat{\beta}-y)+z\big),h_0+h_1'\rangle|\\
&\le&\|F_{T_0\cup T_{10}\cup T_{11}}^T\big((F\hat{\beta}-y)+z\big)\|_2
\|h_0+h_1'\|_2\\
&\le&\sqrt{1.5k}\, 2\lambda \|h_0+h_1'\|_2.
\end{eqnarray*}
\item $ \| \hat{\beta}-\beta\|_2\le \sqrt{2}\big(\|h_0+h_1'\|_2 +
\frac{2 \|\beta_{-\max(k)}\|_1}{\sqrt{k}}\big)$.

In fact,
\begin{eqnarray*}
\| \hat{\beta}-\beta\|_2^2 &=& \|h\|_2^2
=\|h_0+h_1'\|_2^2+\|h_{13}+h_{14}\|_2^2+\sum_{i\ge 2}\|h_i\|_2^2\\
&\le&\|h_0+h_1'\|_2^2+\big(\|h_{13}+h_{14}\|_2+\sum_{i\ge 2}\|h_i\|_2\big)^2\\
&\stackrel{\mbox{by
(\ref{ineq:3.2})}}{\le}&\|h_0+h_1'\|_2^2+\bigg(\|h_0\|_2+
\frac{2 \|\beta_{-\max(k)}\|_1}{\sqrt{k}} \bigg)^2\\
&\le&2\bigg(\|h_0+h_1'\|_2 + \frac{2
\|\beta_{-\max(k)}\|_1}{\sqrt{k}}\bigg)^2.
\end{eqnarray*}
\end{enumerate}
We get the result by combining 1 and 2. This completes the proof.
\qed

We now turn to the second case where the noise $z$ is bounded in
$\ell_2$-norm. Let $F\in \R^{n\times p}$ with $n<p$. The problem is to
recover the sparse signal $\beta\in \R^p$ from
\[
y=F\beta+z
\]
where the noise satisfies $\|z\|_2\le \epsilon.$ We shall again
consider constrained $\ell_1$ minimization:
\[
\min\| \gamma \|_1 \quad \mbox{ subject to } \quad
\|y- F \gamma\|_2 \le \eta.
\]

By using a similar argument, we have the following result.
\begin{thm}\label{thm:4.2}
Let $F\in \R^{n\times p}$. Suppose $\beta\in \R^{p}$ is a $k$-sparse
vector and $y=F\beta +z$  with $\|z\|_{2}\le \epsilon$. If
\begin{equation}
\delta_{1.5k}+\theta_{k, 1.5k} < 1,
\end{equation}
then for any $\eta \ge \epsilon$, the minimizer  $\hat{\beta}$ to
the problem
\[
\min\| \gamma \|_1 \quad \mbox{ subject to } \quad
\|y-F \gamma\|_2 \le \eta
\]
obeys
\begin{equation}
\| \hat{\beta}-\beta\|_2\le C(\eta+\epsilon)
\end{equation}
with $C=\frac{\sqrt{2}(1+\delta_{1.5k})}{1-\delta_{1.5k}-\theta_{k,1.5k}}$.
\end{thm}

\noindent {\bf Proof of Theorem~\ref{thm:4.2} }. Notice that the
condition $\eta \ge \epsilon$ implies that $\|\hat{\beta}\|_1\le
\|\beta\|_1$, so we can use the first part of the proof of
Theorem~\ref{thm:3.2}. The notation used here is the same as that in
the proof of Theorem~\ref{thm:3.2}.

First, we have
\[
\|h_0\|_1 \ge \sum_{i\ge 1}\|h_i\|_1,
\]
and
\[
\|h_0+h_1'\|_2 \le \frac{|\langle Fh, F(h_0+h_1')\rangle|}{\|h_0+h_1'\|_2
\big(1-\delta_{1.5k}-\theta_{k,1.5k}\big)}.
\]

Note that $\|Fh\|_2=\|F(\beta-\hat{\beta})\|_2\le \|F\beta-y\|_2+
\|F\hat{\beta}-y\|_2\le \eta+\epsilon.$

So
\begin{eqnarray*}
\| \hat{\beta}-\beta\|_2 &\le& \sqrt{2}\|h_0+h_1'\|_2\\
&\le & \sqrt{2}\frac{\|Fh\|_2\|F(h_0+h_1')\|_2}{\|h_0+h_1'\|_2
\big(1-\delta_{1.5k}-\theta_{k,1.5k}\big)}\\
&\le &
\sqrt{2}\frac{(\eta+\epsilon)(1+\delta_{1.5k})\|h_0+h_1'\|_2}{\|h_0+h_1'\|_2
\big(1-\delta_{1.5k}-\theta_{k,1.5k}\big)}\\
&\le &\frac{\sqrt{2}(\eta+\epsilon)(1+\delta_{1.5k})}{
1-\delta_{1.5k}-\theta_{k,1.5k}}. \qed
\end{eqnarray*}

\medskip\noindent
{\bf Remarks:} \begin{enumerate} \item Candes, Romberg and Tao
\cite{CanRomTao} showed that, if $\delta_{3k}+3\delta_{4k} < 2$, then
\[
\| \hat{\beta}-\beta\|_2 \le
\frac{4}{\sqrt{3-3\delta_{4k}}-\sqrt{1+\delta_{3k}}}\epsilon.
\]
(The $\eta$ was set to be $\epsilon$ in  \cite{CanRomTao}.)
Now suppose $\delta_{3k}+3\delta_{4k} < 2$. This implies
$\delta_{3k}+\delta_{4k} < 1$ which yields $\delta_{2.4k} +
\theta_{1.6k,2.4k} < 1$, since $\delta_{2.4k}\le \delta_{3k}$ and
$\theta_{1.6k, 2.4k}\le \delta_{4k}$. It then follows from Theorem
\ref{thm:4.2}  that, with $\eta = \epsilon$,
\[
\| \hat{\beta}-\beta\|_2\le
\frac{2\sqrt{2}(1+\delta_{1.5k'})}{1-\delta_{1.5k'}-\theta_{k',1.5k'}}
\epsilon
\]
for all $k'$-sparse vector $\beta$ where $k'=1.6k$.
Therefore Theorem \ref{thm:4.2} improves
the above result in Candes, Romberg and Tao \cite{CanRomTao} by
enlarging the support of $\beta$ by $60\%$.

\item Similar to Theorems \ref{thm:3.2} and \ref{thm:4.1}, we can
have the estimation without assuming that $\hat\beta$ is
$k$-sparse. In the general case, we have
\[
\| \hat{\beta}-\beta\|_2\le
C(\eta+\epsilon)+\frac{2\sqrt{2}\theta_{k,
1.5k}(1-\delta_{1.5k})}{1-\delta_{1.5k}-\theta_{k,1.5k}}k^{-\frac{1}2}\|\beta_{-\max(k)}\|_1.
\]
\end{enumerate}

\subsection*{Connections between RIP and MIP}

In addition to the restricted isometry property (RIP), another commonly
used condition in the sparse recovery literature is the so-called
mutual incoherence property (MIP).
The mutual incoherence property of $F$ requires that the {\sl
coherence bound}
\begin{equation}\label{eq:4-2}
M=\max_{1\le i,j\le p, i\neq j} |\langle f_i, f_j \rangle|
\end{equation}
be small, where $f_1, f_2, \cdots, f_p$ are the columns of
$F$ ($f_i$'s are also assumed to be of length $1$ in $\ell_2$-norm).
Many interesting results on sparse recovery have been obtained by
imposing conditions on the coherence bound $M$ and the sparsity $k$, see
\cite{DET,DonHuo,Fuchs1,Fuchs2,Tropp}. For example, a recent
paper, Donoho, Elad, and Temlyakov \cite{DET}, proved that if
$\beta\in \R^p$ is a $k$-sparse vector and $y=F\beta + z$ with
$\|z\|_2\le \epsilon$, then for any $\eta \ge \epsilon$, the
minimizer  $\hat{\beta}$ to the problem
\[
\min\| \gamma \|_1 \quad \mbox{ subject to } \quad
\|y - F \gamma\|_2 \le \eta
\]
satisfies
\[
\| \hat{\beta}-\beta\|_2\le C(\eta+\epsilon).
\]
with $C=\frac{1}{\sqrt{1-M(4k-1)}}$, provided $k\le \frac{1+M}{4M}$.

We shall now establish some connections between the RIP and MIP and
show that the result of Donoho, Elad, and Temlyakov \cite{DET} can
be improved under the RIP framework, by using Theorem~\ref{thm:4.2}.

The following is a simple result that gives RIP constants from
MIP.

\begin{prop}\label{prop:4.3}
Let $M$ be the coherence bound for $F$. Then
\beq
\delta_{k} \le (k-1)M, \quad \mbox{and } \quad \theta_{k, k'} \le \sqrt{kk'}M.
\eeq
\end{prop}

\noindent {\bf Proof of Proposition~\ref{prop:4.3} }. Let $c$ be a
$k$-sparse vector. Without loss of generality, we assume that
supp$(c)=\{1,2,\cdots, k\}$. A direct calculation shows that
\[
\|Fc\|_2^2 = \sum_{i, j=1}^k \langle f_i, f_j \rangle c_ic_j
=\|c\|_2^2 +\sum_{1\le i, j\le k, i\neq j} \langle f_i, f_j \rangle c_ic_j.
\]
Now let us bound the second term. Note that
\begin{eqnarray*}
\big|\sum_{1\le i, j\le k, i\neq j} \langle f_i, f_j\rangle
c_ic_j\big| &\le&
M\sum_{1\le i, j\le k, i\neq j}|c_ic_j|\\
&\le& M(k-1)\sum_{i=1}^k|c_i|^2 = M(k-1)\|c\|_2^2.
\end{eqnarray*}
These give us
\[
(1-(k-1)M)\|c\|_2^2\le \|Fc\|_2^2 \le (1+(k-1)M)\|c\|_2^2,
\]
and hence
\[
\delta_k \le (k-1)M.
\]

For the second inequality, we notice that $M=\theta_{1,1}$. It then
follows from Proposition~\ref{prop:2.1} that
\[
\theta_{k,k'}\le \sqrt{k'}\theta_{k,1}\le
\sqrt{kk'}\theta_{1,1}=\sqrt{kk'}M. \qed
\]

Now we are able to show the following result.
\begin{thm}\label{thm:4.4}
Suppose $\beta\in \R^{p}$ is a $k$-sparse vector and $y=F\beta +z$
with $z$ satisfying $\|z\|_{2}\le \epsilon$. Let $kM=t$. If
$t < \frac{2+2M}{3+\sqrt{6}}$ (or, equivalently,
$k< \frac{2+2M}{(3+\sqrt{6})M}$), then for any $\eta \ge \epsilon$, the
minimizer  $\hat{\beta}$ to the problem
\[
\min\| \gamma \|_1 \quad \mbox{ subject to }\quad
\|y- F \gamma \|_2 \le \eta
\]
obeys
\begin{equation}
\| \hat{\beta}-\beta\|_2\le C(\eta+\epsilon).
\end{equation}
with $C=\frac{\sqrt{2}(2+ 3t-2M)}{2+2M - (3+\sqrt{6})t}$.
\end{thm}

\medskip\noindent
{\bf Proof of Theorem~\ref{thm:4.4} }. It follows from
Proposition~\ref{prop:4.3} that
\[
\delta_{1.5k}+\theta_{k,1.5k}\le (1.5k+\sqrt{1.5}k-1)M =(1.5+\sqrt{1.5})t-M.
\]
Since $t<\frac{2+2M}{3+\sqrt{6}}$, the condition
$\delta_{1.5k}+\theta_{k, 1.5k} < 1$
holds. By Theorem~\ref{thm:4.2},
\begin{eqnarray*}
\| \hat{\beta}-\beta\|_2 &\le& \frac{\sqrt{2}(1+\delta_{1.5k})}{
1-\delta_{1.5k}-\theta_{k,1.5k}}(\eta+\epsilon)\\
&\le &
\frac{\sqrt{2}(1+ (1.5k-1)M)}{1+M - (1.5+\sqrt{1.5})t}(\eta+\epsilon)\\
&=&\frac{\sqrt{2}(2+ 3t-2M)}{2+2M - (3+\sqrt{6})t}(\eta+\epsilon). \qed
\end{eqnarray*}

\rem In this theorem, the result of Donoho, Elad and Temlyakov
\cite{DET} is improved in the following ways.
\begin{enumerate}
\item The sparsity $k$ is relaxed from $k < \frac{1+M}{4M}$ to
$k < \frac{2+2M}{3+\sqrt{6}M}\approx 1.47\frac{1+M}{4M}$. So roughly
speaking, Theorem \ref{thm:4.4} improves the result in Donoho, Elad
and Temlyakov \cite{DET} by enlarging the support of $\beta$ by 47\%.

\item It is clear that larger $t$ is preferred. Since $M$ is usually
very small, the bound $C$ is tightened from $C=\frac{1}{\sqrt{1+M-4t}}$ to
$C=\frac{\sqrt{2}(2+ 3t-2M)}{2+2M - (3+\sqrt{6})t}$, as $t$ is close to
$\frac{1}4$.
\end{enumerate}

\section{Recovery of Sparse Signals in Gaussian Noise}
\label{sec:Gaussian.noise}

We now turn to the case where the noise is Gaussian. Suppose we
observe
\begin{equation}
\label{Gaussian.model}
y=F \beta + z, \quad z\sim N(0, \sigma^2I_n)
\end{equation}
and wish to recover $\beta$ from $y$ and $F$. We assume that
$\sigma$ is known and that the columns of $F$ are standardized to
have unit $\ell_2$ norm. This is a case of significant interesting, in
particular in statistics. Many methods, including the Lasso
(Tibshirani \cite{Tib}), LARS (Efron, Hastie, Johnstone and
Tibshirani \cite{EHJT}) and Dantzig selector (Candes and Tao
\cite{CanTao07}), have been introduced and studied.

The following results show that, with large probability, the Gaussian
noise $z$ belongs to bounded sets.

\bel
\label{lem:Gaussian.tail}
The Gaussian error $z\sim N(0, \sigma^2I_n)$ satisfies
\beq
P\left(\|F^T z\|_{\infty}\le \sigma \sqrt{2\log p}\right)\ge 1 -
{1\over 2\sqrt{\pi\log p}}
\label{inf.tail}
\eeq
and
\beq
P\left(\|z\|_2 \le \sigma \sqrt{n + 2\sqrt{n\log n}}\right) \ge 1 -
{1\over n}.
\label{l2.tail}
\eeq
\eel
Inequality (\ref{inf.tail}) follows from standard probability
calculations and inequality (\ref{l2.tail}) is proved in the Appendix.

Lemma \ref{lem:Gaussian.tail} suggests that one can apply the
results obtained in the previous section for the bounded error case
to solve the Gaussian noise problem. Candes and Tao \cite{CanTao07}
introduced the Dantzig selector for sparse recovery in the Gaussian
noise setting. Given the observations in (\ref{Gaussian.model}),
the Dantzig selector $\hat \beta^{DS}$ is the minimizer of
\begin{equation}
\label{DS.Gaussian}
(DS)\quad  \min_{\gamma \in \R^p}
\|\gamma\|_1  \quad \mbox{subject to} \quad
\|F^T\big( y - F\gamma \big)\|_{\infty}\le \lambda_p
\end{equation}
where $\lambda_p=\sigma \sqrt{2\log p}$.

In the classical linear
regression problem when $p\le n$ the least squares estimator is the
solution to the normal equation
\begin{equation}
\label{normal.equation}
F^T y = F^T F \beta.
\end{equation}
The constraint  $\|F^T (y - F \beta)\|_\infty \le \lambda_p$ in the
convex program (DS) can thus be viewed as a relaxation of the normal
equation (\ref{normal.equation}). And similar to the noiseless case
$\ell_1$ minimization leads to the ``sparsest'' solution over the
space of all feasible solutions.

Candes and Tao \cite{CanTao07} showed the following result.
\begin{thm}[Candes and Tao \cite{CanTao07}]
\label{thm:2.2} Suppose $\beta\in \R^{p}$ is a $k$-sparse vector
obeying
\[
\delta_{2k}+\theta_{k, 2k}<1.
\]
Choose $\lambda_p=\sigma \sqrt{2\log p}$ in (\ref{eq:1.3}). Then with large
probability, the Dantzig selector $\hat{\beta}$ obeys
\begin{equation}
\| \hat{\beta}-\beta\|_2\le C_1 \sigma \sqrt{k}\sqrt{2\log p},
\end{equation}
with $C_1=\frac{4}{1-\delta_k-\theta_{k, 2k}}$\footnote{It appears
that the constant $C_1$ in Candes and Tao \cite{CanTao07} should be
$C_1=4/(1-\delta_{2k}-\theta_{k, 2k})$.}.
\end{thm}

Another commonly used method in statistics is the Lasso which solves
the $\ell_1$ regularized least squares problem (\ref{lasso}). This is
equivalent to the $\ell_2$-constrained $\ell_1$ minimization problem
($P_1$). In the Gaussian error case, we shall consider a particular
setting. Let $\hat \beta^{\ell_2}$ be the minimizer of
\begin{equation}
\label{l2.Gaussian}
 \min_{\gamma \in \R^p}
\|\gamma\|_1  \quad \mbox{subject to} \quad
\|y - F\gamma\|_2 \le \epsilon_n
\end{equation}
where $\epsilon_n=\sigma \sqrt{n+2 \sqrt{n\log n}}$.

Combining our results from the last section together with Lemma
\ref{lem:Gaussian.tail}, we have the following results on the Dantzig
selector $\hat \beta^{DS}$ and the estimator $\hat \beta^{\ell_2}$
obtained from $\ell_1$ minimization under the
$\ell_2$ constraint. Again, these results improve the previous results in the
literature by weakening the conditions and providing more precise bounds.

\begin{thm}\label{thm:Gaussian}
Suppose $\beta\in \R^{p}$ is a $k$-sparse vector and the matrix $F$ satisfies
\[
\delta_{1.5k}+\theta_{k, 1.5k}<1.
\]
Then with probability $P\ge 1-{1\over 2\sqrt{\pi \log p}}$, the
Dantzig selector $\hat{\beta}^{DS}$ obeys
\begin{equation}
\| \hat{\beta}^{DS}-\beta\|_2\le C_1 \sigma \sqrt{k}\sqrt{2\log p},
\end{equation}
with $C_1=\frac{2\sqrt{3}}{1-\delta_{1.5k}-\theta_{k, 1.5k}}$,
and  with probability at least $1-{1\over n}$, $\hat{\beta}^{\ell_2}$ obeys
\begin{equation}
\| \hat{\beta}^{\ell_2}-\beta\|_2\le D_1 \sigma \sqrt{n+2\sqrt{n\log
n}}
\end{equation}
with
$D_1=\frac{2\sqrt{2}(1+\delta_{1.5k})}{1-\delta_{1.5k}-\theta_{k,1.5k}}$.
\end{thm}

\medskip\noindent
{\bf Remark:} Similar to the results obtained in the previous
sections, if $\beta$ is not necessarily $k$-sparse, in general we
have, with probability $P\ge 1-{1\over 2\sqrt{\pi \log p}}$,
\[
\|\hat{\beta}^{DS}-\beta\|_2\le
C_1 \sigma \sqrt{k}\sqrt{2\log p}+ C_2 k^{-\hf}\|\beta_{-\max(k)}\|_1.
\]
where $C_1=\frac{2\sqrt{3}}{1-\delta_{1.5k}-\theta_{k, 1.5k}}$ and
$C_2 =\frac{2\sqrt{2}(1-\delta_{1.5k})}{1-\delta_{1.5k}-\theta_{k,1.5k}}$,
and with probability $P\ge 1-{1\over n}$,
\[
\| \hat{\beta}^{\ell_2}-\beta\|_2\le
D_1 \sigma \sqrt{n+2\sqrt{n\log n}}+ D_2 k^{-\frac{1}2}\|\beta_{-\max(k)}\|_1
\]
where
$D_1 =\frac{2\sqrt{2}(1+\delta_{1.5k})}{1-\delta_{1.5k}-\theta_{k,1.5k}}$
and $D_2 = \frac{2\sqrt{2}\theta_{k,1.5k}(1-\delta_{1.5k})}{1-\delta_{1.5k}-\theta_{k,1.5k}}$.

\section{Appendix}

{\bf Proof of Proposition \ref{prop:2.3}}. Let
\begin{eqnarray*}
\Lambda &=&\big( (a_1+\cdots+a_w)+2(a_{w+1}+\cdots+a_{2w}) +
(a_{2w+1}+\cdots+a_{3w})\big)^2\\
&=&  \Lambda_1+\Lambda_2+\Lambda_3+\Lambda_4+\Lambda_5+\Lambda_6.
\end{eqnarray*}
Where each $\Lambda_i$ is given (and bounded) by
\begin{eqnarray*}
\Lambda_1&=& \big(a_1+a_2+\cdots+a_w\big)^2\\
&\ge& a_1^2+3a_2^2+\cdots + (2w-1)a_w^2\\
\Lambda_2&=& 4\big(a_{w+1}+a_{w+2}+\cdots+a_{2w}\big)^2\\
&\ge& 4\big(a_{w+1}^2+3a_{w+2}^2+\cdots + (2w-1)a_{2w}^2\big)\\
\Lambda_3&=& \big(a_{2w+1}+a_{2w+2}+\cdots+a_{3w}\big)^2\\
&\ge& a_{2w+1}^2+3a_{2w+2}^2+\cdots + (2w-1)a_{3w}^2\\
\Lambda_4&=& 4\big(a_1+a_2+\cdots+a_w\big)\big(a_{w+1}+a_{w+2}+\cdots+a_{2w}\big)\\
&\ge& 4w\big(a_{w+1}^2+a_{w+2}^2+\cdots + a_{2w}^2\big)\\
\Lambda_5&=& 2\big(a_1+a_2+\cdots+a_w\big)\big(a_{2w+1}+a_{2w+2}+\cdots+a_{3w}\big)\\
&\ge& 2w\big(a_{2w+1}^2+a_{2w+2}^2+\cdots + a_{3w}^2\big)\\
\Lambda_6&=& 4\big(a_{w+1}+a_{w+2}+\cdots+a_{2w}\big)\big(a_{2w+1}+a_{2w+2}+\cdots+a_{3w}\big)\\
&\ge& 4w\big(a_{2w+1}^2+a_{2w+2}^2+\cdots + a_{3w}^2\big).
\end{eqnarray*}
Without loss of generality, we assume that $w$ is even. Write
\[
\Lambda_2=\Lambda_{21}+\Lambda_{22},
\]
where
\[
\Lambda_{21}=4\big(a_{w+1}^2+3a_{w+2}^2+\cdots+
(w-1)a_{w+\frac{w}2}^2+wa_{w+\frac{w}2+1}^2+wa_{w+\frac{w}2+2}^2+\cdots+
wa_{2w}^2\big),
\]
and
\begin{eqnarray*}
\Lambda_{22}&=& 4\big(a_{w+\frac{w}2+1}^2+3a_{w+\frac{w}2+2}^2
\cdots+(w-1)a_{2w}^2\big)
\ge w^2a_{2w}^2\\
&=& (2w-1)a_{2w}^2+(2w-3)a_{2w}^2+\cdots+3a_{2w}^2+\cdots+a_{2w}^2.
\end{eqnarray*}

Now
\begin{eqnarray*}
\Lambda_3+\Lambda_5+\Lambda_6+\Lambda_{22}&\ge&6(w+1)a_{2w+1}^2+(6w+3)a_{2w+2}^2+\cdots
+ (8w-1)a_{3w}^2\\
&& +
(2w-1)a_{2w}^2+(2w-3)a_{2w}^2+\cdots+3a_{2w}^2+\cdots+a_{2w}^2\\
&\ge&6(w+1)a_{2w+1}^2+(6w+3)a_{2w+2}^2+\cdots
+ (8w-1)a_{3w}^2\\
&& +(2w-1)a_{2w+1}^2+(2w-3)a_{2w+2}^2+\cdots+3a_{3w-1}^2+a_{3w}^2\\
&\ge& 8w\big(a_{2w+1}^2+a_{2w+3}^2+\cdots+a_{3w-1}^2+a_{3w}^2\big)
\end{eqnarray*}
and
\begin{eqnarray*}
\Lambda_1+\Lambda_{21}+\Lambda_4
&\ge&a_1^2+3a_2^2+\cdots + (2w-1)a_w^2\\
&& + 4\big(a_{w+1}^2+3a_{w+2}^2+\cdots+
(w-1)a_{w+\frac{w}2}\\
&&+wa_{w+\frac{w}2+1}^2+wa_{w+\frac{w}2+2}^2+\cdots+
wa_{2w}^2\big)\\
&&+4w\big(a_{w+1}^2+a_{w+2}^2+\cdots + a_{2w}^2\big)\\
&\ge&w^2a_w^2
+4(w+1)a_{w+1}^2+4(w+3)a_{w+2}^2+\cdots 4(2w-1)a_{w+\frac{w}2}^2\\
&&+8wa_{w+\frac{w}2+1}^2+8wa_{w+\frac{w}2+2}^2+\cdots+ 8wa_{2w}^2\\
&\ge&\overbrace{4(w-1)a_w^2+4(w-3)a_w^2+\cdots+ 4a_w^2}^{\frac{w}2 \mbox{ terms}}\\
&&+4(w+1)a_{w+1}^2+4(w+3)a_{w+2}^2+\cdots 4(2w-1)a_{w+\frac{w}2}^2\\
&&+8wa_{w+\frac{w}2+1}^2+8wa_{w+\frac{w}2+2}^2+\cdots+ 8wa_{2w}^2\\
&\ge&8w\big(a_{w+1}^2+a_{w+3}^2+\cdots+a_{2w-1}^2+a_{2w}^2\big).
\end{eqnarray*}
Therefore
\[
\Lambda \ge
8w\big(a_{w+1}^2+a_{w+3}^2+\cdots++a_{2w}^2+a_{2w+1}^2+\cdots++a_{3w}^2\big),
\]
and the inequality is proved. \qed

\medskip\noindent
{\bf Proof of Lemma \ref{lem:Gaussian.tail}.} The first inequality
is standard. We now prove inequality (\ref{l2.tail}). Note that
$X=\|z\|_2^2/\sigma^2$ is a $\chi_n^2$ random variable. It follows
from Lemma 4 in Cai \cite{Cai} that for any $\lam > 0$
\[
P(X > (1+\lam) n) \le {1\over \lam \sqrt{\pi n}} \exp\{-{n\over 2} (\lam - \log (1+\lam))\}.
\]
Hence,
\[
P\left(\|z\|_2 \le \sigma \sqrt{n + 2\sqrt{n\log n}}\right)
=1 - P(X > (1+\lam) n)
\ge 1 - {1\over \lam \sqrt{\pi n}} \exp\{-{n\over 2} (\lam - \log (1+\lam))\}
\]
where $\lam = 2\sqrt{n^{-1}\log n}$. It now follows from the fact
$\log (1+\lam)\le \lam - \hf \lam^2 + {1\over 3} \lam^3$ that
\[
P\left(\|z\|_2 \le \sigma \sqrt{n + 2\sqrt{n\log n}}\right) \ge
1 - {1\over n} \cdot {1\over 2 \sqrt{\pi \log n}} \exp\{{4(\log n)^{3/2}\over 3 \sqrt{n}}\}.
\]
Inequality (\ref{l2.tail}) now follows by verifying directly that
${1\over 2\sqrt{\pi \log n}} \exp({4(\log n)^{3/2}\over 3 \sqrt{n}})\le
1$ for all $n \ge 2$. \qed

\end{document}